\documentclass{article}

\usepackage{arxiv}
\usepackage[utf8]{inputenc} 
\usepackage[T1]{fontenc}    
\usepackage{hyperref}       
\usepackage{url}     
\usepackage{nicefrac}       
\usepackage{microtype}      
\usepackage{lipsum}		
\usepackage{graphicx}
\usepackage[numbers]{natbib}
\usepackage{doi}
\usepackage{booktabs, multirow}
\usepackage{float}
\usepackage{marginnote}
\usepackage{fancyhdr}
\usepackage{rotating}
\usepackage{adjustbox}
\usepackage{algorithm}
\usepackage{subcaption}
\usepackage{tabularx}
\usepackage{colortbl}
\usepackage{amsmath,amssymb,amsfonts}
\usepackage{amsthm}
\usepackage{mathrsfs}
\usepackage[dvipsnames]{xcolor}
\usepackage{textcomp}
\usepackage{manyfoot}
\usepackage{algorithmicx}
\usepackage{algpseudocode}
\usepackage{listings}
\usepackage{svg}
\usepackage{pdfpages}
\usepackage{comment}

\title{STOOD-X METHODOLOGY: USING STATISTICAL NONPARAMETRIC TEST FOR OOD DETECTION LARGE-SCALE DATASETS ENHANCED WITH EXPLAINABILITY}

\author{ \href{https://orcid.org/0000-0002-5029-9106}{\includegraphics[scale=0.06]{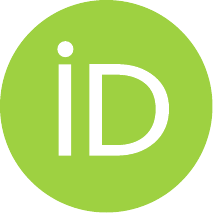}\hspace{1mm}Iván Sevillano-García} \\
	Department of Computer Science and \\
    Artificial Intelligence\\
	Andalusian Research Institute in Data Science and \\Computational Intelligence (DaSCI)\\
	University of Granada, Granada, Spain \\
	\texttt{isevillano@go.ugr.es} \\
	\And
	\href{https://orcid.org/0000-0003-3952-3629}{\includegraphics[scale=0.06]{orcid.pdf}\hspace{1mm}Julián Luengo} \\
	Department of Computer Science and\\
    Artificial Intelligence\\
	Andalusian Research Institute in Data Science and \\Computational Intelligence (DaSCI)\\
	University of Granada, Granada, Spain \\
	\texttt{julianlm@decsai.ugr.es} \\
	  \AND
	\href{https://orcid.org/0000-0002-7283-312X}{\includegraphics[scale=0.06]{orcid.pdf}\hspace{1mm}Francisco Herrera} \\
	Department of Computer Science and\\
    Artificial Intelligence\\
	Andalusian Research Institute in Data Science and \\Computational Intelligence (DaSCI)\\
	University of Granada, Granada, Spain \\
	\texttt{herrera@decsai.ugr.es} \\
}

\begin{document}
\maketitle

\begin{abstract}

Out-of-Distribution (OOD) detection is a critical task in machine learning, particularly in safety-sensitive applications where model failures can have serious consequences. However, current OOD detection methods often suffer from restrictive distributional assumptions, limited scalability, and a lack of interpretability. To address these challenges, we propose \textbf{STOOD-X}, a two-stage methodology that combines a Statistical nonparametric Test for OOD Detection with eXplainability enhancements. In the first stage, STOOD-X uses feature-space distances and a Wilcoxon-Mann-Whitney test to identify OOD samples without assuming a specific feature distribution. In the second stage, it generates user-friendly, concept-based visual explanations that reveal the features driving each decision, aligning with the BLUE XAI paradigm. Through extensive experiments on benchmark datasets and multiple architectures, STOOD-X achieves competitive performance against state-of-the-art post hoc OOD detectors, particularly in high-dimensional and complex settings. In addition, its explainability framework enables human oversight, bias detection, and model debugging, fostering trust and collaboration between humans and AI systems. The STOOD-X methodology therefore offers a robust, explainable, and scalable solution for real-world OOD detection tasks.

\end{abstract}


\section{Introduction}
\label{sec:introduction}

Out-of-Distribution~(OOD) detection has emerged as a challenge within machine learning~\cite{liu2021towards}, which consist in differentiate between In-Distribution~(ID) and OOD samples. In particular, when dealing with Artificial Inteligent~(AI) models, where any instance that can be introduced into the model obtains a prediction, it is essential to recognize when an introduced instance matches the data distribution for which the model has been trained.
In safety-critical scenarios, the absence of OOD algorithms can lead AI models to make incorrect decisions instead of deferring to human judgment.
As a result, the ability to reliably detect OOD samples has become a fundamental requirement for building robust, reliable, and trustworthy AI systems.

Various algorithms have been developed to address this challenge. OOD detection algorithms can be broadly categorized based on where to start applying the algorithm, where we can differentiate training based and post hoc algorithms. Training based algorithms apply their approximation from the training stage. These algorithms modify this stage by adding regularizations to increase separability between ID and OOD samples~\cite{tack2020csi} or even add trainable layers from which to obtain the OOD score~\cite{devries2018learning}. 
Post hoc methods begin their methodology once the model has already been trained. These algorithms are chosen when we have a particular model in production in a real case or the training costs are prohibitive. 

Within the post hoc algorithms, there are different approaches depending on the basis of the algorithm: Classification-based algorithms are based on the model output to detect OOD samples~\cite{liang2017enhancing}. Gradient-based algorithms focus on analyzing the gradients of ID samples to distinguish them from OOD samples~\cite{Park_2023_ICCV}. Distance-based algorithms take advantage of the feature space to detect OOD samples by measuring the distance between ID and OOD samples. 
Some approximations use parametric assumptions such as Gaussianity in the feature space~\cite{lee2018simple,ren2021simple} while others do not, using nonparametric analysis~\cite{pmlr-v162-sun22d}. Furthermore, recent research has explored the combination of the strengths of multiple algorithms~\cite{rajasekaran2024combood}, integrating scores from parametric and nonparametric methods.

Despite these advances, existing OOD detection methods face several limitations. Many approaches rely on strong assumptions about the data distribution, such as Gaussianity, which may not hold in real-world scenarios. Others require computationally expensive procedures or lack explainability, making it difficult to understand how and why a sample is classified as OOD. 
In addition, most of these algorithms propose scores without theoretical robust meaning, such as statistical tests. 
These challenges highlight the need for a more robust and explainable solution for OOD detection.

In parallel to these algorithmic advances, there has been a growing interest in explainable AI~(XAI) techniques that provide insight into how models make decisions~\cite{longo2024explainable}. A well stablished definition on XAI is provided in~\citep{arrieta2020explainable} as "given an audience, an explainable Artificial Intelligence is one that produces details or reasons to make its functioning clear or easy to understand". We can also differentiate between different queries to ask an explainable AI, such as who, when, what, and how to explain a decision~\citep{wang2024roadmap}. A recent deep analysis and reflection on XAI is done in \cite{herrera2025reflections}. Two ways of considering XAI raised in \cite{biecek2024position}, BLUE and RED XAI, bringing the first into a sphere of analysis of expert understanding and trustworthy AI. This first will be used in this paper, considering the distinction between the stakeholders analyzed in \cite{herrera2025reflections} (see Figure 4 of this article, which contains a diagram showing different audience profiles).
Specifically, on the OOD detection task, recent work uses explanations to validate AI decisions~\cite{Dreyer_2024_CVPR}. These approaches take advantage of visualizations to help users understand why a sample is classified as OOD, foster trust, and enable human-AI collaboration. 

In this work, we address the OOD detection problem from a XAI perspective by introducing STOOD-X~(Statistical Test for OOD detection enhanced with eXplainability), a two-stage methodology. 
The first stage of this methodology is a novel post hoc OOD detection algorithm that leverages feature space distances and statistical tests to detect OOD samples.
The second consists in an explanation generation, which provides clear and user-friendly visualization and reasons for decisions made during the OOD detection process.
Unlike some previous approaches, the STOOD-X methodology does not rely on strong distributional assumptions. The STOOD-X methodology uses nonparametric statistical tests to determine whether a new sample belongs to the ID samples with a meaningful score based on statistical probability. 
Additionally, the explanation provided by the second stage enables the final user of the OOD detector to understand the reasons why the STOOD-X methodology makes decisions.

We evaluate the STOOD-X methodology for detection and show the explainability potential of this methodology. 
Our experiments show that the STOOD-X methodology achieves competitive performance compared to state-of-the-art methods on multiple well-known OOD detection datasets using different neural network architectures, particularly in complex, high-dimensional datasets, where the STOOD-X methodology even outperforms the state-of-the-art. These results make the STOOD-X methodology a promising solution for real-world OOD detection tasks. Finally, we include real use cases of the STOOD-X methodology for visualization and show how it enhances explainability, enabling users to understand the model's decision-making process and identify potential biases.

The remainder of this paper is organized as follows. In Section~\ref{sec:related_work}, we review related work in OOD detection and highlight the gaps identified by the STOOD-X methodology. 
Section~\ref{sec:proposal} presents the fundamentals and operation of the STOOD-X methodology for OOD detection, differentiating between the two stages of this methodology and highlighting the importance of each one. 
Section~\ref{sec:experimental_setup} describes the experimental setup of different scenarios, while Section~\ref{sec:experimental_results} shows the results, including comparisons with state-of-the-art methods. Section~\ref{sec:visualization} shows how the STOOD-X methodology visualizations enhance the explainability, enabling users to understand the model's decisions and identify potential biases. Finally, Section~\ref{sec:conclusions} concludes the paper and discusses future research directions. 
\section{Related Works}
\label{sec:related_work}

In this section, we present a comprehensive review of the literature on OOD detection, focusing on both theoretical advancements and practical applications.
In Section~\ref{sec:ood_detection_related_work} we begin with a discussion of theoretical proposals and improvements for OOD detection. We differentiate the different types of algorithms by their principles and methodologies.
Then, in Section~\ref{sec:xai_related_work}, we explore how XAI approaches are used to bring the OOD decision to human understanding.

\subsection{OOD Detection algorithms}
\label{sec:ood_detection_related_work}

In this section, we analyze the different approaches that have been used to tackle the problem of OOD detection.
Specifically, here is a brief summary of the different OOD proposals on which the STOOD-X methodology has been inspired.
For a more detailed study of the taxonomy of generalized OOD methods, we refer to~\cite{yang2024generalized}.

\begin{figure}
    \centering
    \includegraphics[trim=0cm 22cm 0cm 0cm, clip, width=0.7\textwidth]{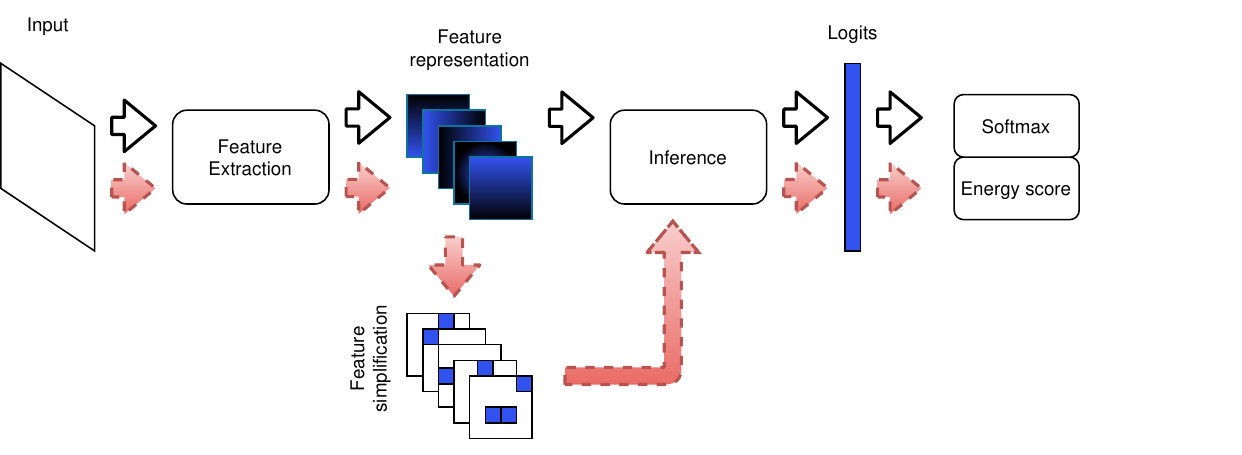}
    \caption{Diagrammatic representation of the ASH methodology for estimating OOD confidence via Energy scores of features simplification.}
    \label{fig:ash-based-figure}
\end{figure}

We can differentiate between OOD detection algorithms between training-based and post hoc methods. 
Training based algorithms are developed to impose a set of constraints in training time so that the resulting model performs better on the OOD detection task. An example of this training-based algorithm can be found in~\cite{devries2018learning}, where a modified model is trained to estimate its confidence as a scalar in $[0,1]$. 
Other approximations of this training-based perspective impose regularization factors, such as contrastive learning~\cite{tack2020csi}, on the training stage to facilitate separability between OOD and ID samples.
Post hoc algorithms study the OOD detection problem for an already trained AI model, without modifying the training stage.
This property is important in a real-world environment where the cost of retraining a model with new OOD restrictions is prohibitive. 
An example of this approximation is ODIN~\cite{liang2017enhancing}, which uses temperature scaling and input perturbation to improve the separability between ID and OOD samples.
ODIN's approach to amplifying differences in softmax scores has inspired subsequent research, including methods that use energy scores~\cite{liu2020energy} for the detection of OOD. Algorithms such as ASH~\cite{djurisic2022extremely}, ReAct~\cite{sun2021react}, ViM~\cite{wang2022vim} or TempScaling~\cite{xu2024scaling} use the energy score obtained from logits by simplifying the internal representation of features to maximize the difference between the ID and OOD samples. In Figure~\ref{fig:ash-based-figure}, we show graphically how the ASH algorithm simplifies the internal representation of the features of the model to infer whether the behavior of the model on a new sample is an OOD or ID sample based on the energy score of the modified representation. The white arrows represent the natural flow of the prediction model. The red arrows represent the flow used by ASH to calculate the OOD confidence.

Other algorithms base their OOD scores on gradients instead of logits or feature representation. This is the case with GradNorm~\cite{huang2021importance}, which uses the magnitude of gradients as an OOD detector.
These methods focus on improving the discriminative power of OOD scores by taking advantage of different aspects of the model output. However, recent works~\cite{igoe2022useful} demonstrate that the success of gradient-based OOD detectors does not necessarily depend on the gradient itself but on the magnitude of the learned features.

More recently, distance-based methods have emerged as a promising direction in OOD detection. These methods rely on the assumption that OOD samples are relatively far from ID classes in the feature space. 
This intuition of these methods appears naturally, as the model learns a feature space where ID sample classes are distinguishable from each other. However, if an OOD sample is brought into the feature space, it may have features of different classes anecdotally, although not all the features required to belong to a particular class, nor as present as a sample of the class itself.

These methods usually work in the same way: the distance from the ID distribution is calculated. If this distance exceeds a certain threshold, it is considered an OOD sample. The search of this threshold is done experimentally and is optimized to separate ID and OOD data.
However, this threshold is empirically searched and evaluated by the experimental result, without a theoretical robustness to support the results. 

There are distance-based algorithms with very different principles. Parametric algorithms such as the minimum Mahalanobis distance score~(MDS)~\cite{lee2018simple}, or its variation Relative-MDS(RMDS)~\cite{ren2021simple}, study the distances from the OOD sample to the centroid of each class assuming that the features are normally distributed. Alternatively, there are nonparametric algorithms that make no strong assumptions about the underlying feature distribution, making it a versatile and effective method. The work proposed in~\cite{Park_2023_ICCV} demonstrated the effectiveness of nonparametric K-nearest-neighbor (KNN) distances as a nonparametric OOD detection method. Finally, a last avenue explored is the combination of parametric and nonparametric algorithms. This is the case with CombOOD~\cite{rajasekaran2024combood}, which blends the RMDS and KNN scores to obtain a combined OOD score.

Although distance-based algorithms can use various distances, recent work~\cite{techapanurak2020hyperparameter,chen2020boundary} shows that the cosine distance between features is useful for differentiation between OOD and ID examples. This distance works especially well when the vectors to be measured are sparse, with many dimensions with a value of 0. 

\subsection{XAI in OOD Detection}
\label{sec:xai_related_work}

OOD detection algorithms aim to detect OOD samples and differentiate them from ID samples. However, once detected, these algorithms must provide reasons why these samples have been detected as OOD. This is where XAI is introduced to propose human-understandable reasoning so that humans can evaluate whether each sample is OOD. 

Within the field of study of XAI, two perspectives can be distinguished~\citep{biecek2024position}. 
The RED XAI~(Research, Explore, Debug XAI) refers to a field of explainability focused on the development and resolution of AI while the BLUE XAI~(responsiBle, Legal, trUst, Ethics) aims to improve the proposal of explanations to a final user, ensuring its good behavior. Within both approaches, ensuring that an OOD detection algorithm is understood by a final user is part of the BLUE XAI perspective.

Answering the question of how to present an explanation, we can distinguish between several explanation proposals~\citep{wang2024roadmap}: Explanations based on the importance of the feature or explanations based on examples. On the one hand, explanations based on feature importance assign a percentage of influence to each part of the sample input, showing how much it has influenced the final decision. It is presented to the final user by a heatmap highlighting these importances. However, explanations based on examples show samples similar in some sense to the intended example, either to enforce the decision (prototypes) or to show the changes needed to change these decisions (counterfactual).

Explanations based on feature importance have two different perspectives, black-box and white-box. Black-box algorithms do not use the internal structure of a model~\cite{ribeiro2016should} so that the explanation is not biased by the specifications of the model itself. However, these explanations require a large number of model evaluations to provide a reliable explanation.
The white box perspective takes advantage of the knowledge of the internal structure to develop more specific explanations, such as Layer-wise Relevance Propagation(LRP)~\cite{montavon2019layer}, which preserves the importance of the decision along the internal layers of the model. Moreover, by being able to use the knowledge of the internal structure, these algorithms can propose explanations in reasonable time and cost.

Based on LRP, Concept-based Relevance Propagation~(CRP)\cite{achtibat2023attribution} has been developed. CRP introduces the term 'concept' to separate different features within the same explanation to obtain GLocal explanations(Global-Local). GLocal explanations offers a local perspective~(where the concept is located on the example to explain) and a global perspetive~(which examples has the same concepts present as the example to explain). 

Based on CRP and making use of prototypes, Prototypical Concept-based Explanations~(PCX)~\cite{Dreyer_2024_CVPR} is proposed, an explanation proposal for different tasks, including the OOD detection task. For the OOD detection approximation, this proposal can be classified as a parametric distance-based OOD detector, which computes the distances to class prototypes.
However, the main contribution of this work in the XAI perspective is the explanation proposal, which presents prototypical dataset examples identified by the AI model, highlighting similar features to the analyzed sample and their locations. 
This method bases its explanation proposal on the selection of prototypes. For a method that does not use prototypes to detect OOD examples, the explanation must be adapted to align with the underlying approach to remain meaningful.

\section{STOOD-X methodology: OOD detection algorithm using feature space analysis and statistical tests enhanced by explainability}
\label{sec:proposal}
 
In this section, we describe the fundamentals of the STOOD-X methodology, a novel explainable two-stage methodology designed to detect OOD samples using feature space distances to Nearest Neighbors~(NNs) and statistical tests with improved explainability. Due to its feature-based construction of observed ID samples, it provides an approximation that can contribute to the explainability of the OOD detector. 
This methodology ensures a unified framework that combines accurate OOD detection with robust explainability, empowering stakeholders to make informed decisions while maintaining trust in the AI system.

We organize the description of the STOOD-X methodology as follows.
We begin with Section~\ref{sec:flowchart}, which outlines the overall workflow of the STOOD-X methodology, providing a comprehensive perspective on its integrated approach.
In Section~\ref{sec:formulation}, we formulate the mathematical fundamentals and rationales for the first stage of the STOOD-X methodology algorithm proposal. 
In Section~\ref{sec:XAI_formulation}, we detail how the second stage of the STOOD-X methodology takes advantage of the formulation of the STOOD-X methodology to develop explanations based on the importance of the NN characteristics compared to the important features of the new sample with a perspective of BLUE XAI.

\subsection{Flowchart of the STOOD-X methodology}
\label{sec:flowchart}

\begin{figure}[ht]
    \centering
    \includegraphics[width=.9\linewidth,trim=0cm 14.5cm 0cm 0cm, clip]{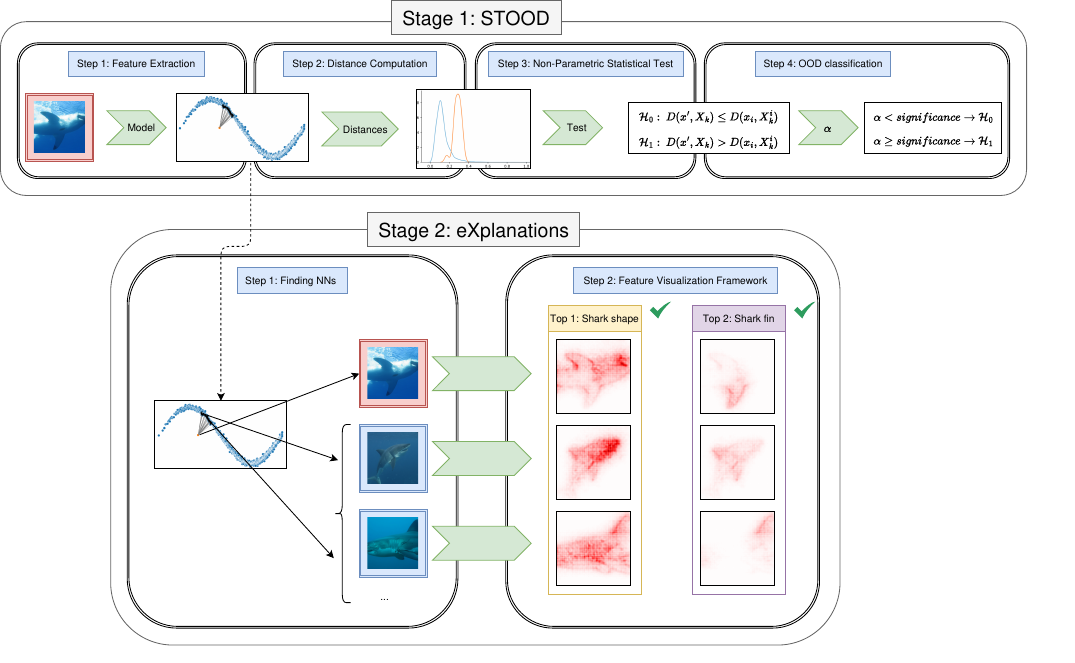}
    \caption{Flowchart of the STOOD-X methodology}
    \label{fig:flowchart}
\end{figure}

In this section, we describe the natural workflow of the STOOD-X methodology, from the moment a new sample is presented to the model until it determines whether the sample is ID and processes it accordingly, or alternatively, identifies it as an OOD sample and requests user validation with an explanation. Figure~\ref{fig:flowchart} illustrates this workflow, emphasizing how statistical analysis and explainability mechanisms interact synergistically to provide reliable detection and human-understandable justifications.

The STOOD-X methodology executes through two sequential yet interdependent stages, each with distinct operational objectives:
\begin{enumerate}
    \item In Stage 1, the original model processes its features and maps them to the learned feature space. Within this space, the distances to previously observed samples are analyzed and a statistical hypothesis test is performed to determine whether the new sample is sufficiently close to the validated ID samples. Based on a predefined significance threshold, the detection algorithm classifies the new sample as ID or OOD. Two possible outcomes follow:
    \begin{itemize}
        \item If the new sample is classified as ID, it is deemed ready to be processed by the model, as there are validated nearby samples.
        \item If the sample is classified as OOD, it should be excluded from processing, as no similar instances exist in the feature space.
    \end{itemize}

    \item Stage 2 of the STOOD-X becomes relevant, providing additional insights into the model's decision-making process by presenting nearby samples along with the most important features shared between the new sample and its closest counterparts. The explanation generated is valuable for the understanding of the model of the final user, regardless of whether the sample is classified as ID or OOD.
    \begin{itemize}
        \item In the ID case, the user should be able to verify that the features used by the model align with those of validated samples.
        \item In the OOD case, the final user should also validate that the identified features do not provide meaningful or comparable information relative to previously validated samples.
    \end{itemize}
    
    Although we have already mentioned that Stage 2 is useful in both cases, the explanations generated are particularly valuable in the case of OOD classification scenario, the model’s uncertainty can be effectively translated into a meaningful query for the final user. By highlighting the closest validated samples, the model implicitly asks whether the new sample should be processed despite not having encountered a similar instance before. This enables the user to assess a validation of the model's decision and take appropriate action. As a result, the approach helps prevent OOD samples from being mistakenly processed as ID while also allowing the reconsideration of ID samples that were misclassified as OOD, ensuring a more reliable decision-making process.
\end{enumerate}

In the next sections, we will go into the specifics of both stages. These detailed explanations will provide a clearer understanding of the processes involved in each stage, building on the general description given earlier.

\subsection{STOOD-X methodology first stage: OOD detection algorithm}
\label{sec:formulation}

In this section, we introduce the OOD detection algorithm that constitutes the first stage of the STOOD-X methodology. We begin by formally defining the key concepts and notation that establish the basis of our approach, ensuring a clear and rigorous foundation. This formalization provides the necessary framework to systematically derive the STOOD-X methodology, allowing a principled approach to OOD detection.

Let $\mathcal{X}, \mathcal{V}, \mathcal{Y}$ be the input, feature, and output space, respectively. Let $f:\mathcal{X} \rightarrow \mathcal{Y}$ be a machine learning model. 
This model $f$ can be modeled as follows:
\begin{multline}
\\V: \mathcal{X} \rightarrow \mathcal{V}  = \mathbb{R}^d \\
C: \mathcal{V} \rightarrow \mathcal{Y} \\
f = C \circ V: \mathcal{X} \rightarrow \mathcal{Y},\\
\end{multline}
where $V$ is the feature extractor and $C$ is the inference function from the features.

\begin{figure}[ht]
    \centering
    \includegraphics[trim=0cm 21cm 3cm 0cm, clip,width=0.7\linewidth]{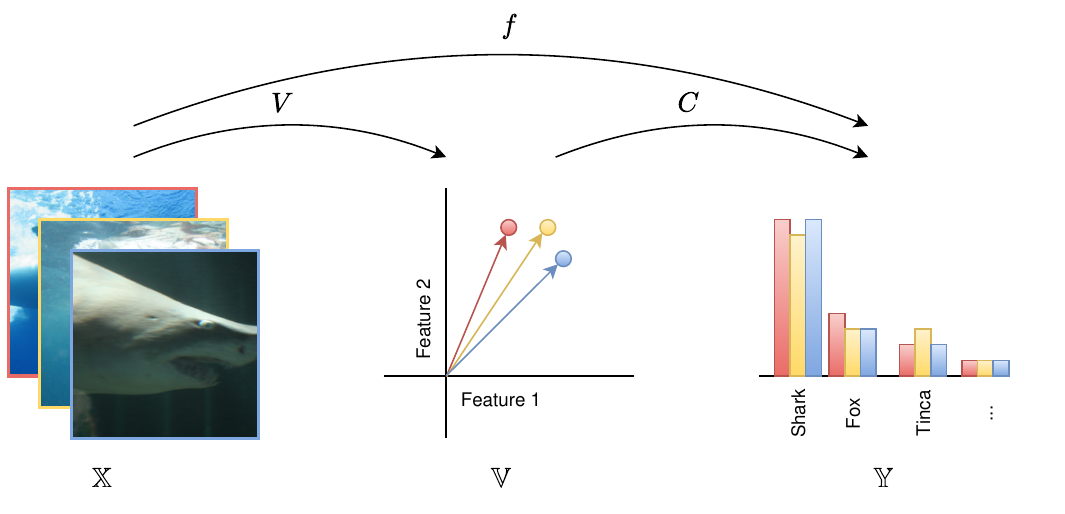}
    \caption{Representation of the machine learning models into two separated functions $V$ and $C$}
    \label{fig:f_representation}
\end{figure}

This representation is always possible, as $V$ and $C$ can correspond to the function $f$ and the identity, or the other way around. For the representation of $f$ as a machine learning model, it is useful and straightforward to separate the two parts. 
The reason is that machine learning models usually have a feature extractor part and an inference part that computes these features for the desired output. 
In Figure~\ref{fig:f_representation}, we show the representation of a classification model with two main features in which the classification ideally obtains similar results and whose features are concentrated in the same feature space region. 

The model $f$ has been trained and tested in all $(\mathbb{X}_{train}, \mathbb{Y}_{train})$ and $(\mathbb{X}_{test}, \mathbb{Y}_{test})$, respectively. 
Intuitively, the train and test sets are samples of the same distribution in $\mathcal{X}$ and, since $f$ has been trained with these data, the features extracted with $V$ are also of the same distribution. In a classification problem, we may consider each class as a different random variable. In our approximation, the features of a sample $x$, $V(x)$, will be near other samples of the same class and will be separated from the features of other classes. Moreover, in the case of an OOD sample, its features will be separated from all the samples features of the original classes. 

\begin{figure}[b]
    \centering
    \begin{subfigure}[b]{0.3\textwidth}
        \centering
        \includegraphics[width=\linewidth]{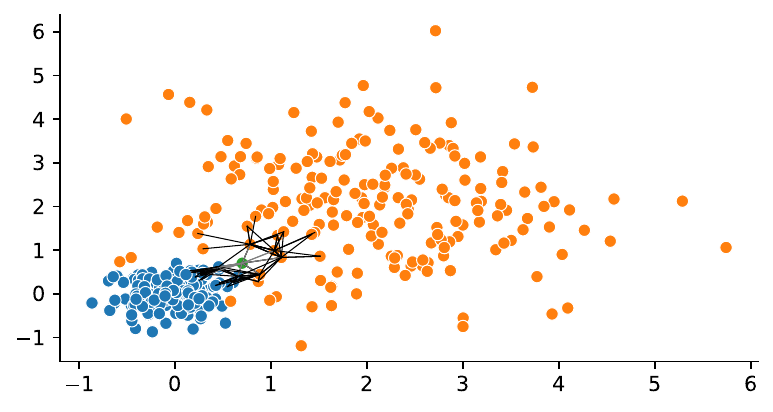}
        \caption{Feature distribution}
        \label{fig:subfig_distribution1}
    \end{subfigure}
    \begin{subfigure}[b]{0.3\textwidth}
        \centering
        \includegraphics[width=\linewidth]{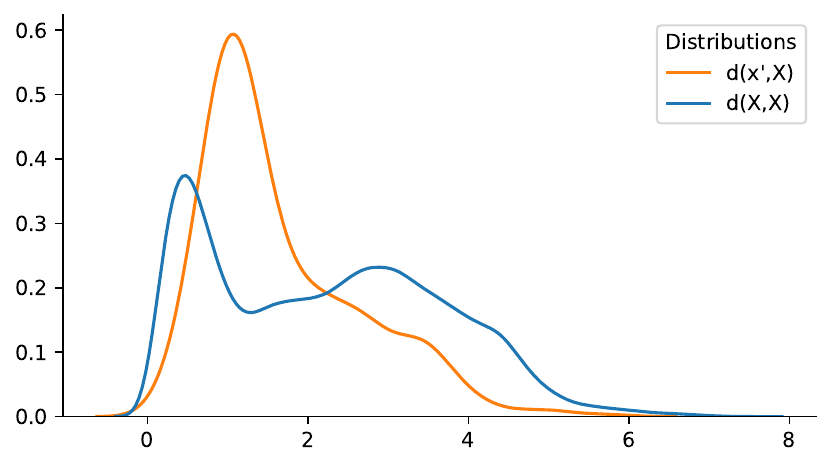}
        \caption{Distances distribution}
        \label{fig:subfig_distribution2}
    \end{subfigure}
    \begin{subfigure}[b]{0.3\textwidth}
        \centering
        \includegraphics[width=\linewidth]{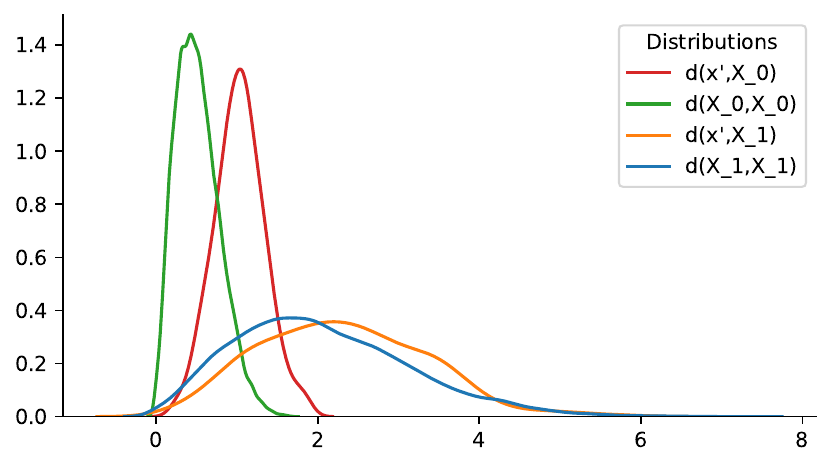}
        \caption{Distances distributions per class}
        \label{fig:subfig_distribution3}
    \end{subfigure}
    \caption{Intuition of the behavior in the feature space of ID~(orange and blue) and OOD~(green) samples}
    \label{fig:id_ood_samples}
\end{figure}

It is worth noting that in classification tasks we can differentiate between several distributions, one for each class. In Figure~\ref{fig:id_ood_samples}, we show the distributions of features and distances in the feature space of two different classes in the feature space. We can model this scenario with several approaches:
\begin{itemize}
    \item \textbf{Distribution of distances to samples in $X_{set}$.} This case is the most general. We consider that the whole dataset is part of the same distribution. However, we do not exploit the a priori knowledge concerning the class that the model attributes. In Figure~\ref{fig:subfig_distribution2}, we show the distribution of the distances of the OOD sample compared to the distribution of the ID samples. We differentiate between the two distributions, but we would not be able to determine whether one distribution lies above or below the other.
    
    \item \textbf{Distribution of distances to each sample in $X_{set}$ separated by class.} This case considers distance distributions differently per class. The ID set membership depends on whether the sample $x'$ belongs to one of these distributions. In Figure~\ref{fig:subfig_distribution3}, we show the difference between the distances of the OOD sample $x'$ to each class compared to the distances of each class. In this example, we may observe that the distances from $x'$ to class 0 are slightly larger than they might be, but the distances to class 1 are quite similar.
    
    \item \textbf{Distribution of distances corresponding to the samples in $X_{set}$ whose class is the one selected by the model.} Looking at the distributions in the figure, we can differentiate between the distances to class 0 and class 1. If a model were to classify this OOD sample as class 0, its ID membership could be low because its distances to samples of class 0 are larger than usual. However, the distances to class 1 are equivalent, so one could classify this example as ID with the above approach, even with its misclassification. Therefore, from a present viewpoint the approach is to use only the class that the model has predicted.
\end{itemize}

The STOOD-X methodology has two main advantages: it allows us to check whether a sample is OOD or not, and it also allows us to differentiate between samples that are on the decision boundary.
These samples, although they belong to the ID distribution, are samples in which the model is not truly confident and may give erroneous inferences. Therefore, we base our proposal on the final prediction class of the model.

Based on the previous argument, and specializing the proposal in distinguishing whether a sample is ID or OOD of an unique distribution, we formalize the following approach:
Let $x', x_1, ..., x_N$ samples in $\mathcal{X}$ where $x',x_i$ are samples of the same distribution and $X_{set}=\{x_1,...,x_N\}$ be samples of the distribution $\mathbb{X}$. Then, for a distance $d:V \times V \rightarrow [0,\inf)$, we compute the distances of $V(x')$ and $V(x_i)$ for $i\in X_{set}$, that is, $d(V(x'),V(x_i))$. 
Without loss of generality, we can assume that $x_i$ are sorted with respect to the distance $d(V(x'),V(x_i))$ in ascending order. For $x'$ and $1 \leq k \leq N$, we can define $X_{k}={x_1,...,x_k}$ the k NNs to $x'$ with distance $D:\mathbb{X} \times \mathbb{X} \rightarrow [0,\inf) ,D(z_1,z_2) = d(V(z_1),V(z_2))$.
Also, for each $x_i \in X_k$ we can define $X^i_k={x^i_1,...,x^i_k}$ the $k$ NNs of $x_i$ in $X_{set} \setminus \{x_i\}$.
Both $X_k$ and $X^i_k$ are the k NNs of $x'$ and $x_i$, respectively. 

\begin{figure}[ht]
    \centering
    \begin{subfigure}[b]{0.45\textwidth}
        \centering
        \includegraphics[width=\linewidth]{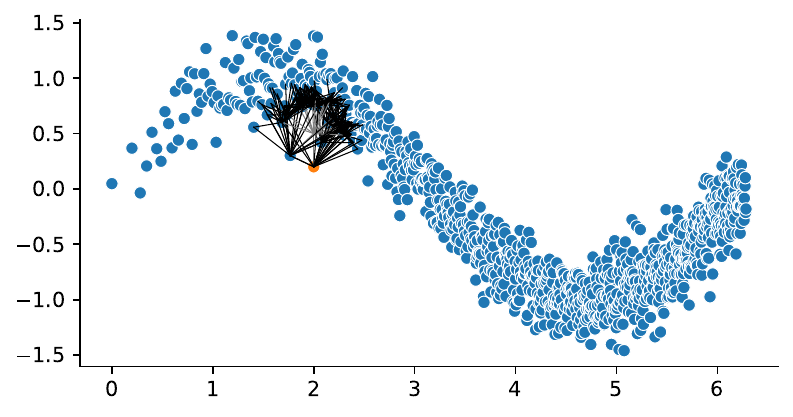}
        \caption{Feature distribution}
        \label{fig:subfig_distribution1_sin}
    \end{subfigure}
    \begin{subfigure}[b]{0.45\textwidth}
        \centering
        \includegraphics[width=\linewidth]{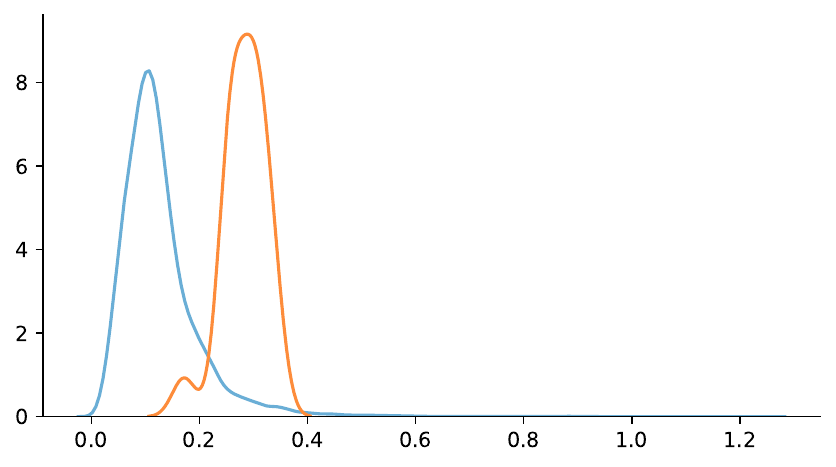}
        \caption{Distances distribution}
        \label{fig:subfig_distribution2_sin}
    \end{subfigure}
    \caption{Feature samples and distances distribution of $(x,sin(x)+\mathcal{N}(0,0.2)$ (blue) and an OOD sample(orange). In gray, the connection on OOD sample and its NNs. In black, the connections of the OOD's NNs and their NNs.}
    \label{fig:feat_dist_distr_sin}
\end{figure}

For illustration purposes, in Figure~\ref{fig:feat_dist_distr_sin} we show an example in which the ID data set is distributed as $(x,sin(x)+\mathcal{N}(0,0.2)$~(blue) where we have introduced the OOD sample $(2,0)$~(orange). In Figure~\ref{fig:subfig_distribution1_sin}, we notice that most NNs in the OOD sample have their own NNs at a smaller distance on average than the NNs of the OOD sample $x'$. In Figure~\ref{fig:subfig_distribution2_sin}, we confirm this point by showing the distribution of the distances of the different NNs (blue) and the OOD sample (orange). 

The main hypothesis of the STOOD-X methodology consists of the following assumption: If $x'$ belongs to the original distribution, there will be no significant differences between the distance distributions $D(x',X_{k})$ and $D(x_i,X^i_k)$. As we assume $x'$ and $x_i$ are samples of the same distribution, the set of distances $D(x',X_{k})$ and $D(x_i,X^i_k)$ are also samples of the same distribution: \emph{"Distance on the feature space of a sample of X to their k NNs"}. Otherwise, if $x'$ does not belong to the original distribution, the distances $D(x',X_{k})$ should be greater than the distances $D(x_i,X^i_k)$. This translates into the specific case of Figure~\ref{fig:subfig_distribution2} as we should differentiate whether the OOD has a larger distance to samples from the original distribution than its NN.

The distribution of features does not necessarily follow any concrete distribution, so the set of distances $D(x',X_{k})$ and $D(x_i,X^i_k)$ also does not necessarily follow any particular distribution.  
To be able to statistically differentiate between both distributions, we use the nonparametric Wilcoxon-Mann-Whitney test for the differences of two variables in its positive version. This test takes as null hypothesis that the random variable $X-Y \leq 0$ and as an alternative hypothesis, $X-Y > 0$. 
The p-value obtained by this test will be the score that the STOOD-X methodology assigns to the new sample. This is a percentage that indicates how likely the new sample is to belong to the original distribution, and by selecting a specific significance, it can be determined whether the sample is ID or OOD.

As shown in Figure~\ref{fig:flowchart}, the first stage of the STOOD-X methodology can be summarized as follows in 4 steps:
\begin{enumerate}
    \item \textbf{Feature extraction:} Extract the features of the new sample $x'$ with the feature extractor of our model.
    \item \textbf{Distance computation:} Compute the distances of the new sample $x'$ to the samples in the train set $X_{set}$ with the distance function $D$ on the feature space.
    \item \textbf{Nonparametric Statistical Test:} Apply the Wilcoxon-Mann-Whitney test to the distances of the new sample $x'$ to the NNs $X_{k}$ and the distances of the NNs $x_i$ to their NNs $X^i_k$.
    \item \textbf{OOD classification:} Depending on the significance level, classify the new sample $x'$ as ID or OOD. The significance level indicates the confidence with which we want to classify a new sample as ID.
\end{enumerate}

In summary, this first stage of the STOOD-X methodology provides a principled method for OOD detection by using feature-space distances and statistical tests. Its ability to handle nonparametric distance distributions and its adaptability to various machine learning models make it a versatile tool to improve the reliability of AI systems.

\subsection{STOOD-X second stage: eXplanation generation}
\label{sec:XAI_formulation}

In order to justify the decisions made by the STOOD-X methodology in a friendly way to a final user, it should be studied how this methodology can explain their decisions from a BLUE XAI perspective. Initially, the p-value used as score is a percentage value with respect to how far the new sample is from the train distribution, but the quality of the explanation to a final user can be increased through different types and explainability techniques.

As an example-based explanation, we can show $NNs$ within the class so that a user can evaluate whether the neighbors have similar features. The STOOD-X methodology score is calculated based on how far the new sample is from the set of trains in the feature space. Therefore, showing the NNs is a good explanation of the features that make the algorithm increase or decrease the score. 

As an explanation based on the importance of the features, we can show the locations of crucial details for the decision of the model on the new sample. Based on algorithms such as PCX, we can also analyze the presence of concepts studied a priori within the new sample, distilling the final decision of the algorithm into separate features analysis of the new sample.

By the nature of the STOOD-X methodology, the combination of both perspectives can provide more explanation for OOD detection. We can show the local explanations separately per feature of the new sample while we show the explanations of the NNs. In this way, the final user can qualitatively evaluate whether the explanations offered for each feature are similar to those offered for the NNs.

To justify the OOD score obtained, the STOOD-X methodology proposes to combine the explanations of importance of the features of the new sample and the NN of the new example explanations, thus categorizing this explanation as a BLUE XAI proposal.
As shown in Figure~\ref{fig:flowchart}, the second stage of the STOOD-X methodology is based on the following two steps:

\begin{enumerate}
    \item \textbf{Finding NNs in the feature space:} We select the train samples with the closest features in the feature space of the STOOD-X methodology formulation. Since the STOOD-X methodology score is based on the distances to train samples on the feature space, the closest samples must have similar visualizations if classified as ID.
    The final user will be able to establish relationships within those samples and checking whether the closest samples have similar visualizations, thus justifying the score proposed by the STOOD-X methodology first stage.
    \item \textbf{Feature Visualization Framework:} For the most present features in these train samples, the STOOD-X methodology visualizations show their feature importance localization from the train samples and the new sample. This step is essential to confirm whether specific features of the new sample correspond to biases or to features present in the original dataset. It is in this visualization where we can establish whether we validate the features learned by our model or instead whether it introduces a bias that should be avoided in the model.
\end{enumerate}

It should be noted that this explanatory stage is not always required. While explanations can be generated on demand, they become particularly valuable in cases of uncertainty, either when the first stage identifies a clear OOD sample or when confidence in the ID classification is low. In these scenarios, the STOOD-X methodology presents its explanations as an open question to the final user, highlighting features that the algorithm recognizes as novel or uncertain. This human validation step becomes crucial when the system encounters unfamiliar patterns, effectively making the user a collaborator in resolving ambiguous cases. Thus, the STOOD-X methodology detects distributional shifts while actively engaging humans in the decision-making process for borderline samples.

As we show, the explanation offered is faithful to the behavior of the STOOD-X methodology for detection. Moreover, it takes advantage of its design to provide useful information for its understanding. In short, the second stage of the STOOD-X methodology is designed to propose explanations in line with the BLUE XAI perspective and enhance human-machine cooperation.

\section{Experimental setup}
\label{sec:experimental_setup}

In this section, we provide a detailed description of the experimental setup used to evaluate the OOD detection stage of the STOOD-X methodology performance in the OOD detection task.
Specifically, we describe the selection of benchmarks~\ref{sec:benchmark}, the neural network architectures selected for each scenario~\ref{sec:arquitecture}, the performance metrics employed for the evaluation~\ref{sec:metrics_performance} and the hyperparameters studied for the STOOD-X methodology fine-tuning~\ref{sec:hyperparameters_studied}. Finally, in Section~\ref{sec:sota_algorithms}, we detail the state of the art OOD algorithms with which the STOOD-X methodology we compare.

\subsection{Benchmark selection}
\label{sec:benchmark}

To evaluate the performance of our method, we use the OpenOOD framework~\cite{zhang2023openood}, which provides a variety of ID datasets along with the corresponding OOD datasets. For each ID dataset, OpenOOD distinguishes between Near and Far OOD datasets based on the degree of similarity between the ID and OOD datasets. We show the specifications of each experimentation scenario in Table~\ref{tab:benchmark_table}.

\begin{table}[ht]
    \centering
    \begin{tabular}{lccc}\toprule
ID dataset &Near OOD &Far OOD \\\midrule
Cifar10~\cite{cifar10} &CIFAR-100~\cite{cifar100}, ImageNet200~\cite{wu2017tiny} &MNIST~\cite{deng2012mnist}, SVHN~\cite{netzer2011reading}, Textures~\cite{kylberg2011kylberg}, Places365~\cite{zhou2017places} \\ \midrule
Cifar100~\cite{cifar100}  &CIFAR-10~\cite{cifar10}, ImageNet200~\cite{wu2017tiny} &MNIST~\cite{deng2012mnist}, SVHN~\cite{netzer2011reading}, Textures~\cite{kylberg2011kylberg}, Places365~\cite{zhou2017places} \\ \midrule
Imagenet200~\cite{wu2017tiny}  &SSB-hard~\cite{tajwar2021no}, NINCO~\cite{bitterwolf2023or} & iNaturalist~\cite{huang2021mos}, Textures~\cite{kylberg2011kylberg}, OpenImage-O~\cite{wang2022vim}  \\ \midrule
Imagenet1K~\cite{krizhevsky2012imagenet}  & SSB-hard~\cite{tajwar2021no}, NINCO~\cite{bitterwolf2023or} &iNaturalist~\cite{huang2021mos}, Textures~\cite{kylberg2011kylberg}, OpenImage-O~\cite{wang2022vim} \\
\bottomrule
\end{tabular}     \caption{Specifications of the benchmark scenarios provided by OpenOOD}
    \label{tab:benchmark_table}
\end{table}

\subsection{Neural Network Architectures selection}
\label{sec:arquitecture}

The experimental setup involves testing various models, depending on the ID dataset. For the selected benchmarks and post hoc methods to be compared, specific models and weights have been commonly used in the literature for the correct comparison between OOD detection methods. 

For CIFAR-10, CIFAR-100, and ImageNet200, we use ResNet18 architecture~\cite{he2016deep} with pre-trained checkpoints provided by OpenOOD. For the sake of fair comparison with existing proposals, experiments are carried out on these datasets with the ResNet18 architecture, since comparative experiments of previous work are based on this architecture.

For the larger ImageNet dataset, we employ two distinct architectures: ResNet50~\cite{he2016deep} and ViT-B/16)~\cite{dosovitskiy2020image}. Both models use pre-trained checkpoints available in torchvision~\cite{marcel2010torchvision}. The inclusion of these models allows us to evaluate the performance of our method across different architectural paradigms: convolutional neural network and vision transformers. In addition, these architectures are two of the architectures widely used for comparing other state-of-the-art works. 

\subsection{Metrics for OOD Detection Performance Evaluation}
\label{sec:metrics_performance}
To perform the performance analysis, we evaluated our method using the metrics most commonly used for OOD detection. We mainly use the Area Under the Receiver Operating Characteristic curve~(AUROC) and the False Positive Rate under a True Positive Rate of 95\%~(FPR@95).
The implementation of both metrics can be found in the OpenOOD framework~\cite{yang2022openood}.

The AUROC metric is the standard metric reported on the OpenOOD online leaderboard. AUROC provides a comprehensive measure of the model's ability to separate ID and OOD samples across all possible classification thresholds. For this reason, AUROC is a robust and widely accepted metric for OOD detection tasks. It is important to note that the AUROC metric is better when higher, with a value of 1 indicating perfect separation between ID and OOD samples, and a value of 0.5 indicating random performance. Additionally, AUROC is threshold independent, which ensures that the evaluation is not biased by the choice of a specific decision boundary. 

FPR@95 is particularly useful for understanding the model's behavior in high-recall scenarios, which are critical in real-world applications where minimizing false positives is essential. It also helps us in providing a more complete picture of the model's performance. FPR@95 is better when it is lower, as a lower value indicates that the model can achieve a high true positive rate while keeping false positives to a minimum.

\subsection{STOOD-X methodology hyperparameters}
\label{sec:hyperparameters_studied}
The design decisions for the STOOD-X methodology are guided by the goal of achieving a balance between performance and computational efficiency. In the following, we outline the key hyperparameters and the rationale behind their selection.

\begin{itemize}
    \item \textbf{Distance Metric (d)}: We use the cosine distance as a distance metric for the feature space. The cosine distance is particularly suitable for high-dimensional spaces, such as the feature space.
    \item \textbf{Neighbor Set}: For the set of possible neighbors, we use the training subset. This choice gives us a large number of possible neighbors without introducing data snooping into the experiment.
    \item \textbf{Number of Neighbors ($K$)}: We first analyze the influence of $K$, the number of neighbors considered, by testing values such as 9, 18, 36, 72, 144, 288, 500, and $N$ (the size of the entire training set). Our analysis reveals that increasing $K$ improves detection performance, but with diminishing returns beyond a certain point. Based on this trade-off between performance and computation time, we set $K=500$ for subsequent experiments.
    \item \textbf{Number of important Features ($N_f$)}: We also analyze the influence of $N_f$ for distance computation. We tested various percentages of features, including 12.5\%, 25\%, 37.5\%, 50\%, 62.5\%, 75\%, 87.5\%, and 100\%.
\end{itemize}

\subsection{State of the art OOD algorithm}
\label{sec:sota_algorithms}
To assess the performance of the STOOD-X methodology for detection, we compare it with state-of-the-art OOD detection algorithms. Since the STOOD-X methodology is a post hoc OOD detection approach, we ensure a fair evaluation by selecting competing methods from the same category and applying them to the same pre-trained model. Specifically, we consider post hoc algorithms that rank among the top two performers in at least one dataset within the OpenOOD framework's OOD detection leaderboard.

As discussed in Section~\ref{sec:related_work}, post hoc algorithms encompass various approaches. The selected algorithms cover a diverse set of these perspectives, ensuring a comprehensive comparison. Table~\ref{tab:sota_descr} provides an overview of the underlying perspectives for each algorithm included in our evaluation.

\begin{table}[ht]
    \centering
    \scriptsize
\begin{tabular}{lcccccc}
\toprule
\textbf{Method} & \textbf{Distance-Based} & \textbf{Energy-Based} & \textbf{Gradient-Based} & \textbf{Perturbation} & \textbf{Prediction-Based} \\
\midrule
ASH~\cite{djurisic2022extremely}        &        &       Yes               &                      & Yes (features)  &   \\
CombOOD~\cite{rajasekaran2024combood}    & Yes  &                      &                      &                                        &   \\
Gradrm~\cite{huang2021importance}       &                          &                      & Yes (input gradients) &                                        &   \\
KNN~\cite{pmlr-v162-sun22d}               & Yes        &                      &                      &                                        &   \\
MDS~\cite{lee2018simple}                  & Yes        &                      &                      &                                        &   \\
NNGuide~\cite{Park_2023_ICCV}             & Yes        &                      & Yes (auxiliary)       &                                        &   \\
ODIN~\cite{liang2017enhancing}             &                          &                      & Yes (for perturbation) & Yes (gradient)                         & Yes \\
ReAct~\cite{sun2021react}                  &                          &                      &                      & Yes                         &   \\
RMDS~\cite{ren2021simple}                 & Yes  &  &                      &                                        &   \\
TempScaling~\cite{xu2024scaling}          &                          &                      &                      &                                        & Yes \\
VIM~\cite{wang2022vim}                    &                          & Yes                  &                      &                                        &   \\
\bottomrule
\end{tabular}     \caption{Categorization of state of the art post hoc OOD Detection algorithms}
    \label{tab:sota_descr}
\end{table} 

\section{Evaluating the STOOD-X methodology capabilities in OOD detection}
\label{sec:experimental_results}

This section analyzes the behavior of the first stage of the STOOD-X methodology, which involves the OOD detection algorithm. The analysis is carried out in three steps, two to empirically optimize the STOOD-X methodology and one to evaluate its performance.. In Section~\ref{sec:nn}, we analyze the influence of the number of neighbors used. In Section~\ref{sec:nf}, we evaluate the impact of the number of features used to calculate the distances. Finally, in Section~\ref{sec:sota}, we compare the performance of the optimized STOOD-X methodology with other state-of-the-art algorithms for OOD detection.

\subsection{Influence of the number of neighbors \texorpdfstring{$K$}{K}}
\label{sec:nn}

In this section, we investigate the impact of varying the number of neighbors, $K$, on the performance of the STOOD-X methodology for detection. By testing different values of $K$ in the Imagenet dataset with ViT-B16 architecture, we aim to determine the optimal number of NNs that balance model performance with computational efficiency. 
The results are presented in Table~\ref{tab:nn}, where we compare different $K$ values in various performance metrics.

\begin{table}[ht]
    \centering
    \begin{tabular}{lrrrrrr}\toprule
$K$ &AUROC Near ($\uparrow$) &FPR@95 Near($\downarrow$) &AUROC Far($\uparrow$) &FPR@95 Far($\downarrow$) &Time(s in cpu) \\\midrule
9 &72.328 &100 &74.837 &100 &4 \\
18 &74.571 &85.428 &77.831 &76.269 &4 \\
36 &76.545 &74.859 &80.791 &68.719 &5 \\
72 &78.545 &71.747 &85.498 &52.618 &6 \\
144 &80.217 &67.328 &88.443 &41.972 &10 \\
288 &81.252 &63.954 &90.31 &35.855 &19 \\
500 &81.681 &\textbf{62.001} &91.196 &33.213 &30 \\
5000 &\textbf{81.899} &62.11 &\textbf{92.181} &\textbf{30.167} &269 \\
\bottomrule
\end{tabular}     \caption{Comparison of the number of neighbors~($K$) in the Imagenet dataset with the ViT-B16 arquitecture in terms of different metrics of the STOOD-X methodology}.
    \label{tab:nn}
\end{table}

Upon reviewing the performance metrics, we observe a clear trend in the relationship between the number of neighbors and the model's performance in the AUROC and FPR@95 metrics. As $K$ increases, both metrics show a noticeable improvement in the detection of near-and far-OOD samples. This suggests that considering a larger number of neighbors helps the model better capture the underlying structure of the data, thereby enhancing its ability to distinguish between ID and OOD samples. 
However, as highlighted in the time column of Table~\ref{tab:nn}, the computational time increases significantly with larger values of $K$. This is expected as evaluating a greater number of neighbors requires more calculations, which directly impacts the execution time of the method.

Provided that there is a trade-off between performance and computational efficiency, we conducted a detailed analysis to find an optimal balance. Although increasing $K$ results in higher performance, the improvement decreases after a certain threshold. 
For example, the jump in quality from $K=500$ to $K=5000$ the maximum number of NNs is relatively small compared to the large increase in execution time. Based on these observations, we decided to select $K=500$ for the rest of our experiments. This choice strikes a reasonable balance, offering satisfactory performance with a manageable computational cost, avoiding excessive time consumption for minimal gains in quality.

\subsection{Influence of the percentage of Features considered}
\label{sec:nf}

In this section, we explore the impact of the number of features selected on the performance of the STOOD-X methodology, specifically analyzing how excluding less important features affects detection accuracy. We evaluate the STOOD-X methodology using different percentages of features and compare its performance across several datasets provided by OpenOOD. Furthermore, we test two different architectures for ImageNet: ResNet50 and ViT-B16, to examine how the number of features influences both neural network structures. These architectures are chosen because they are commonly used in the OpenOOD framework, providing a standardized basis for comparison. We show the results in Table~\ref{tab:NF}. 

\begin{table}[ht]
    \centering
    \scriptsize
\begin{tabular}{l|rr|rr|rr|rr|rr}\toprule
&\multicolumn{2}{c}{Cifar10: 95,22\%} &\multicolumn{2}{c}{Cifar100: 77.17\%} &\multicolumn{2}{c}{Imagenet200: 86.38\%} &\multicolumn{2}{c}{Imagenet (R50): 80.38\%} &\multicolumn{2}{c}{Imagenet (ViT): 81.14\%} \\\cmidrule{2-11}
Nº Features &Near &Far &Near &Far &Near &Far &Near &Far &Near &Far \\\midrule
100.0\% &\textbf{89.527} &\textbf{91.884} &\textbf{80.363} &81.179 &82.004 &\textbf{90.538} &\textbf{77.66} &85.308 &81.51 &90.742 \\
87.5\% &89.520 &91.853 &80.347 &81.165 &82.048 &90.527 &77.652 &85.304 &81.521 &90.806 \\
75.0\% &89.433 &91.772 &80.255 &\textbf{81.215} &82.076 &90.454 &77.649 &85.301 &81.548 &90.892 \\
62.5\% &89.267 &91.553 &79.99 &81.079 &82.151 &90.397 &77.647 &85.294 &81.576 &90.947 \\
50.0\% &88.833 &90.912 &79.662 &80.929 &\textbf{82.172} &90.29 &77.649 &85.294 &81.63 &91.065 \\
37.5\% &88.305 &89.981 &79.211 &80.301 &82.16 &90.112 &\textbf{77.66} &\textbf{85.311} &81.681 &91.196 \\
25.0\% &88.118 &89.531 &78.728 &79.1 &82.104 &89.943 &77.65 &85.264 &\textbf{81.695} &\textbf{91.241} \\
12.5\% &87.926 &89.197 &77.597 &77.946 &81.822 &89.687 &77.538 &84.989 &81.574 &91.118 \\
\bottomrule
\end{tabular}     \caption{Performance of the STOOD-X methodology on the datasets provided in OpenOOD with different percentage of features used}
    \label{tab:NF}
\end{table}

From the results on the CIFAR-10 and CIFAR-100 datasets using the ResNet18 architecture, we observe a consistent trend: reducing the number of features used to calculate the distance between neighbors leads to a deterioration in performance. 
This suggests that the feature space for each class is dispersed across various dimensions, and removing any of these dimensions negatively impacts the model's ability to detect OOD examples. In particular, performance does not degrade significantly when only 12.5\% of the most important features are excluded, implying that the remaining 87.5\% of the features considered less important do not substantially affect the distinction between ID and OOD samples.

However, the performance on ImageNet200 (using ResNet18), as well as on ImageNet with the ResNet50 and ViT-B16 architectures, presents an interesting countertrend: reducing the number of features slightly enhances performance. This result is initially counterintuitive, as the removal of features would typically lead to a loss of information, as in the previous experiment. However, it suggests that many of the features in these neural networks are redundant or even detrimental to the OOD detection task. Specifically, the less important features seem to introduce noise, negatively impacting the model's ability to differentiate between ID and OOD examples.

This finding also has notable implications from an XAI perspective. By reducing the number of features considered, the complexity of the model is reduced, making it easier to explain the decisions made by the OOD detector. This can be particularly valuable when understanding the rationale behind the model’s decisions is crucial.

Based on these observations, we select the optimal percentage of features for each dataset, which corresponds to the configuration that yields the best performance in Near AUROC for each architecture and dataset.

\subsection{Performance analysis}
\label{sec:sota}

Once the STOOD-X methodology behavior has been analyzed, we compare it against several state-of-the-art algorithms for OOD detection. The primary metric used for the evaluation is the AUROC, which is calculated for near- and far- OOD scenarios. The OpenOOD library was used to obtain experimental results for state-of-the-art algorithms whenever possible, ensuring consistency in the evaluation process. For methods not integrated within OpenOOD, we carefully implemented the algorithms based on the details provided in their respective publications to maintain a fair and consistent comparison.

For the STOOD-X methodology, the optimal configuration of each hyperparameter (number of neighbors $K$ and percentage of features) was selected based on the previous sections: From~\ref{sec:nn}, we choose $K=500$ as a balance in performance and computational time, and from~\ref{sec:nf}, we choose the best number of features considered for each architecture and dataset based on the highest AUROC achieved in the near-OOD scenario.

\begin{table}[ht]
    \centering
    \scriptsize
\begin{tabular}{lrr|rr|rr|rr|rrr}\toprule
&\multicolumn{2}{c}{Cifar10} &\multicolumn{2}{c}{Cifar100} &\multicolumn{2}{c}{Imagenet200} &\multicolumn{2}{c}{Imagenet Resnet50} &\multicolumn{2}{c}{Imagenet ViT B16} \\\cmidrule{2-11}
Method &Near &Far &Near &Far &Near &Far &Near &Far &Near &Far \\\midrule

ASH~\cite{djurisic2022extremely} &74.111 &78.360 &78.394 &79.701 &82.119 &94.226 &36.312 &30.469 &53.206 &51.555 \\
CombOOD~\cite{rajasekaran2024combood}* &90.101 &92.759 &80.708 &\textbf{82.947} &83.348 &90.519 &80.512 &88.072 &79.601 & 92.652 \\
GradNorm~\cite{huang2021importance} &53.772 &58.553 &69.734 &68.816 &73.327 &85.293 &38.817 &32.636 &39.281 &41.746 \\
KNN~\cite{pmlr-v162-sun22d} &\textbf{90.699} &93.105 &80.248 &82.317 &81.750 &93.474 &70.100 &88.640 &74.112 &90.812 \\
MDS~\cite{lee2018simple} &86.716 &90.201 &58.794 &70.062 &62.507 &74.939 &76.038 &\textbf{93.473} &79.042 &92.599 \\
NNGuide~\cite{Park_2023_ICCV} &52.261 &46.820 &77.089 &76.357 &76.150 &90.683 &38.765 &53.355 &40.906 &54.387 \\
ODIN~\cite{liang2017enhancing} &80.253 &87.210 &79.798 &79.440 &80.320 &91.897 &67.944 &67.970 &64.306 &76.058 \\
React~\cite{sun2021react} &86.468 &91.019 &80.705 &79.844 &80.484 &93.096 &36.142 &36.227 &69.261 &85.687 \\
RMDS~\cite{ren2021simple} &89.534 &92.427 &80.27 &82.528 &82.904 &88.54 &\textbf{80.612} &87.535 &80.088 &92.6 \\
TempScaling~\cite{xu2024scaling} &82.215 &87.906 &\textbf{80.94} &81.421 &\textbf{85.114} &\textbf{94.307} &67.741 &75.543 &58.896 &75.037 \\
VIM~\cite{wang2022vim} &88.506 &\textbf{93.136} &74.833 &82.114 &78.814 &91.52 &64.542 &92.112 &77.029 &\textbf{92.837} \\ \midrule
\textbf{STOOD-X} &89.527 &92.013 &80.363 &81.215 &82.172 &90.538 &77.660 &85.311 &\textbf{81.948} &92.198 \\
\bottomrule
\end{tabular}     \caption{Near and Far AUROC metric of the STOOD-X methodology}. *Implemented by our team, due to the absence of a suitable pre-existing OpenOOD implementation
    \label{tab:sota_comparison}
\end{table}

The proposed method demonstrates robust and competitive performance in all datasets evaluated, consistently ranking among the top performing algorithms. The results summarized in Table~\ref{tab:sota_comparison} highlight that the STOOD-X methodology are on par with or superior to several state-of-the-art approaches. Here is a breakdown of the results by dataset and architecture:

\begin{itemize}
    \item \textbf{CIFAR10}: The STOOD-X methodology achieves AUROC scores of 89.527\% in the Near OOD scenario and 92.013\% in the Far OOD scenario. These scores are competitive with the best methods such as KNN (90.699\%, 93.105\%) and CombOOD (90.101\%, 92.759\%). The results indicate that the STOOD-X methodology can effectively handle near- and far- OOD scenarios, making it a robust option for a variety of detection scenarios. Other algorithms, such as ViM with the highest Far AUROC~(93,136\%) has a worse performance on Near AUROC~(88.506\%) realizing that there is a trade-off between Far and Near scenarios.

    \item \textbf{CIFAR100}: In the Near OOD scenario, the STOOD-X methodology achieves an AUROC of 80.363\%, slightly below top methods like TemScaling (80.94\%) and CombOOD (80.708\%). In the Far OOD scenario, the STOOD-X methodology reaches 81.215\%, which is competitive, but still slightly behind methods like CombOOD~(82.947\%) and KNN~(82.317\%). These results indicate that the method performs well in this dataset, although there is room for improvement compared to the leading methods. On this dataset we notice again a tradeoff between Near and Far scenarios.

    \item \textbf{Imagenet200}: In the Far OOD scenario, the STOOD-X methodology achieves an AUROC of 90.538\%, outperforming methods such as RMDS (88.54\%) but far from the top performing TempScale (94.307\%). In the Near OOD scenario, it reaches 82.172\%, which performs also worse than TempScale~(85.114\%). These results underline the method's capability to handle complex datasets, demonstrating its strength in high-dimensional spaces with more diverse and challenging OOD examples, but with further improvements.

    \item \textbf{ImageNet (ResNet50 architecture)}: In Far scenarios, the STOOD-X methodology performs with an 85.294\%, behind VIM (92.112\%) and MDS (93.473\%). However, when we compare with the Near scenarios, the performance achieved by the STOOD-X methodology~(77,
    .66\%) outperforms both methods~(ViM: 64.542\%, MDS:76.038\%). RMDS(80.612\% Near, 87.535\% Far) and CombOOD(80.512\% Near, 88,077\% Far) are examples of balance between Near and Far AUROC, with worse performance in the Far scenario but outperforming imbalanced algorithms in the Near scenario. The STOOD-X methodology is competitive compared to the balanced algorithms.

    \item \textbf{ImageNet (ViT-B16 architecture)}: The STOOD-X methodology achieves 81.948\% (Near) and 92.198\% (Far), outperforming  state-of-the-art balanced algorithms, such as RMDS with 80.088\% and 92.6\% in the Near and Far scenarios, respectively. Imbalanced algorithms like ViM~(77.029\% Near, 92.837\% Far) or KNN~(74.112\% Near, 90.812\% Far) could achive a good result in the Far scenarios with a worse performance in the Near. The results suggest that the STOOD-X methodology is suitable for modern architectures such as transformers, which are gaining prominence in computer vision tasks.
\end{itemize}

In conclusion, the STOOD-X methodology shows strong and consistent performance in multiple datasets, maintaining a balance between near- and far-reach scenarios. 
Although it is competitive with the best performing algorithms, there are specific areas where it can be further optimized, particularly for convolutional networks. 
The STOOD-X methodology excels in more complex detection scenarios, especially with transformer-based models, indicating its versatility and adaptability. 
Additionally, the number of classes in each dataset may have influenced the quality of the OOD detector, as larger class sets can present more challenging scenarios for detection. 
With further refinement, particularly in feature extraction and architectural adjustments, the proposed method has the potential to be a leading solution for OOD detection in a variety of computer vision tasks. 
\section{Evaluating the STOOD-X methodology explainability}
\label{sec:visualization}

In this section, we show how the STOOD-X methodology can provide explainability with a BLUE XAI perspective to present explanations to the final users. For this purpose, we use the zennit-crp library~\cite{achtibat2023attribution} to visualize the importance of features in images.

Two different final-user examples are discussed below. These cases correspond to a test ID case and an OOD case with a low confidence score. As mentioned in Section~\ref{sec:XAI_formulation}, low-confidence cases occur when the algorithm does not find nearby examples validated by humans. In such scenarios, the explanation serves as a prompt for the end user to verify the decision, actively involving them in the process for ambiguous cases.

\begin{figure}
    \centering
    \includegraphics[trim=0cm 12cm 0cm 0cm, clip, width=0.7\textwidth]{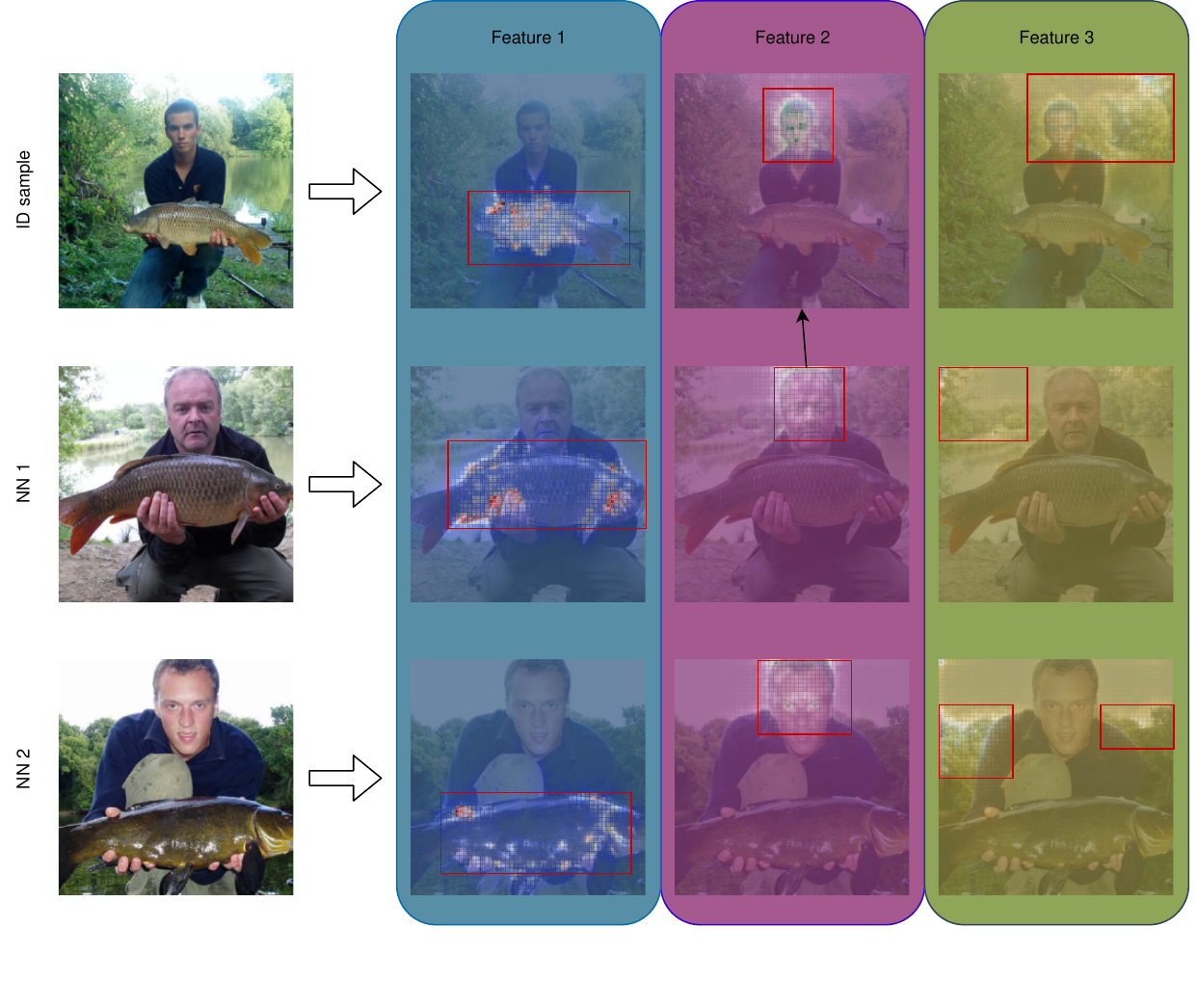}
    \caption{Example of OOD study results shown to an end user of an ID example with an 52\% ID score}
    \label{fig:tinca}
\end{figure}

In Figure~\ref{fig:tinca}, we show an example of the first Imagenet class, a Tinca fish, identified as an ID test sample with membership in ID 52\%. By choosing the two NNs, we can see concrete similarities to the new sample. Looking at the three most present features, we see that both neighbors and the test sample recognize similar patterns per feature. 
The first feature focuses attention on the fish, validating its importance for this class. The second feature focuses their attention on the man's head. This feature is used to detect a biasing pattern for the classification of the fish: the presence of a person holding the fish. For further study, this feature should be ignored. The third feature is focused on the background, so it should also be ignored for the fish classification. 
Moreover, a final user could propose to introduce this example as part of the ID set to enhance STOOD-X methodology.

\begin{figure}
    \centering
    \includegraphics[trim=0cm 12cm 0cm 0cm, clip, width=0.7\textwidth]{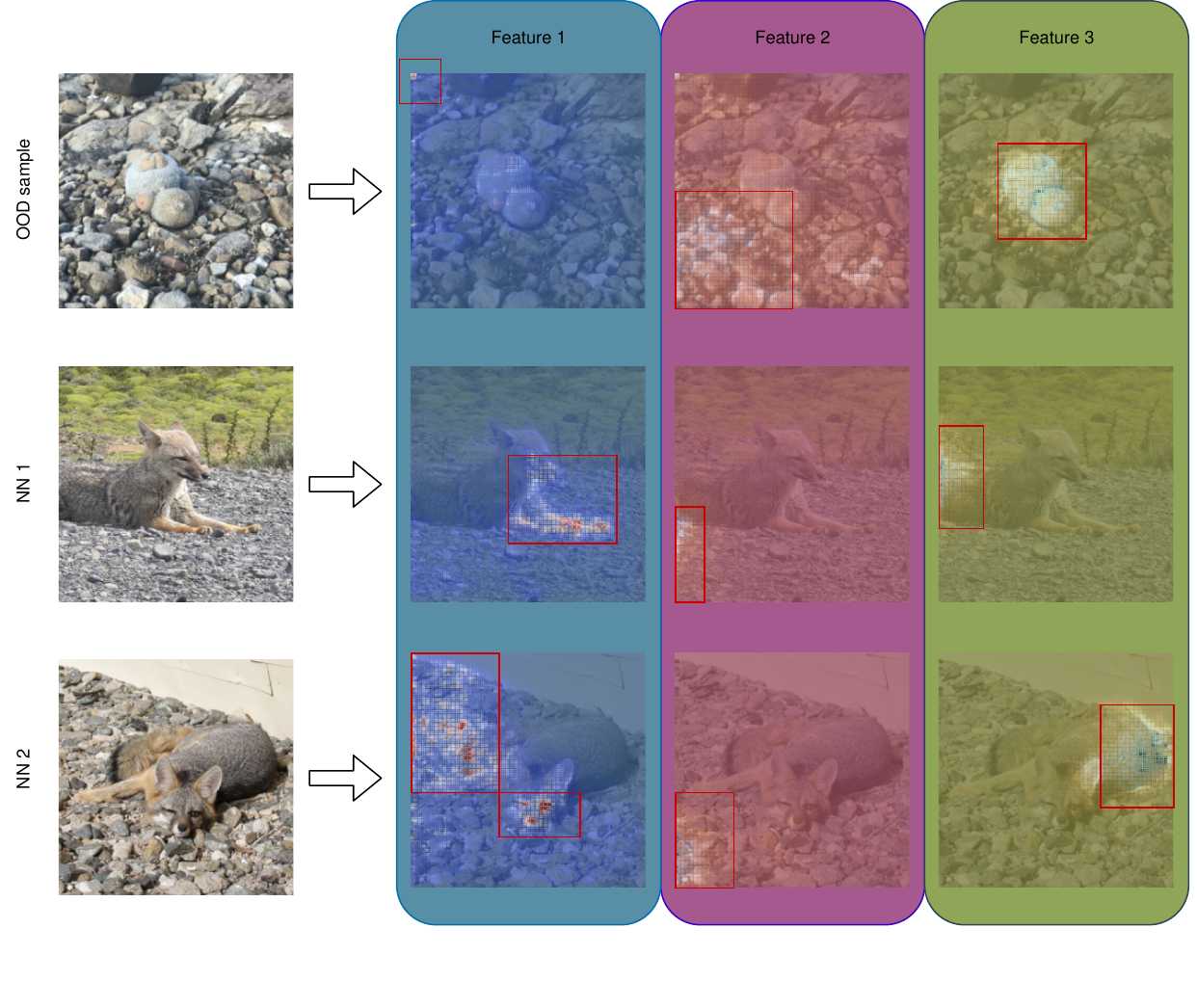}
    \caption{Example of OOD study results shown to an end user of an OOD example with a 2\% ID score}
    \label{fig:ood_ninco}
\end{figure}

In Figure~\ref{fig:ood_ninco}, we show an example of the NINCO dataset classified as fox. In this example, the model has assigned a very low ID membership score of 2\%, indicating a high likelihood that the sample is outside the known distribution. 
The analysis of NNs reveals no similarities to the typical features of foxes in the original example. However, upon closer inspection of the most important features, the model seems to have focused on detecting a fox face, but it instead highlights a central stone, which is unrelated to the fox class. Furthermore, the second most relevant feature is the presence of stones in the background, which is also found in the neighbors. The third most present feature seems to be related to the back of the fox, also detected on the central stone.
This observation of the second feature points out that the model has unintentionally learned a bias toward background features (the stones) rather than the actual fox features.
The final user, upon reviewing these findings, should recognize that the model is relying on irrelevant background cues for classification and report that these biases are skewing the decision-making process. 
This could prompt efforts to delete these from the model, ensuring that it focuses more accurately on the relevant animal features.

This analysis highlights the importance of providing explanations for the STOOD-X methodology. 
By providing clear and detailed visualizations of key features, users can better understand how and why the model classifies an example in a particular way. 
This approach also helps identify potential areas for improvement or biases. For example, in cases with low ID membership (e.g., the fox sample~\ref{fig:ood_ninco}), irrelevant features such as stones in the background can lead to misclassification, indicating the need to disregard such biases. 
On the other hand, ID samples (e.g., the tinca fish sample~\ref{fig:tinca}) can validate decisions while also detecting biases in the model, allowing the model to explain. 
Therefore, incorporating visual explanations of the STOOD-X methodology separated by features allows users to make more informed decisions and refine the behavior of the model. 

In summary, the STOOD-X methodology stands out for its ability to bridge the gap between algorithmic decision making and human comprehension, fostering effective human-AI collaboration. The STOOD-X methodology empowers users to understand with its explanations and refine OOD classifications, enhancing trust and facilitating AI integration in critical domains. It can be used to detect biases and analyze system capabilities, improving the system reliability and enabling continuous model improvements. Furthermore, the STOOD-X methodology promotes seamless human-machine collaboration through intuitive visualizations, ensuring that AI-driven decisions align with human expertise. 
\section{Conclusions}
\label{sec:conclusions}

The STOOD-X methodology presents a significant advance in OOD detection by combining robust statistical analysis with human-centered explainability. Its two-stage framework, which uses nonparametric statistical tests for detection and explainable visualizations for decision support, offers a principled, scalable, and user-interpretable solution. Unlike many existing methods, the STOOD-X methodology does not rely on restrictive distributional assumptions and provides statistically meaningful confidence scores through the Wilcoxon-Mann-Whitney test.

Empirical results across diverse benchmarks, including CIFAR and ImageNet datasets, demonstrate the competitive performance of the STOOD-X methodology compared to state-of-the-art post hoc OOD detectors. The method consistently balances performance across both near- and far-OOD scenarios while maintaining computational efficiency. Moreover, its alignment with the BLUE XAI perspective enhances trust and transparency, offering meaningful concept-driven visual explanations that help uncover both strengths and biases in model behavior.

Beyond technical contributions, the STOOD-X methodology highlights the importance of a trustworthy and explainable AI in safety-critical domains such as healthcare, finance, and autonomous systems. Its ability to detect and visualize decision-relevant features empowers domain experts to audit and improve model behavior.

Future directions include extending the STOOD-X methodology to other modalities (e.g. time series), integrating with active learning pipelines, and advancing user interfaces for richer interaction with explanations. With its foundation in statistical rigor and human-centered design, the STOOD-X methodology sets a promising path for reliable, transparent, and adaptive AI systems in real-world environments. 
\section*{Acknowledgments}

F. Herrera is supported by the TSI-100927-2023-1 Project, funded by the Recovery, Transformation, and Resilience Plan from the European Union Next Generation through the Ministry for Digital Transformation and the Civil Service.
All authors also receive support from the Spanish Ministry of Science and Technology under project PID2023-150070NB-I00 financed by MCIN/AEI/10.13039/501100011033. 

\bibliographystyle{unsrtnat}
\bibliography{references}

\end{document}